# REVIEWING DATA ACCESS PATTERNS AND COMPUTATIONAL REDUNDANCY FOR MACHINE LEARNING ALGORITHMS

Imen Chakroun, Tom Vander Aa and Tom Ashby
*ExaScience Life Lab, IMEC, Leuven, Belgium*
{imen.chakroun, tom.vanderaa, tom.ashby}@imec.be

**ABSTRACT**

Machine learning (ML) is probably the first and foremost used technique to deal with the size and complexity of the new generation of data. In this paper, we analyze one of the means to increase the performances of ML algorithms which is exploiting data locality. Data locality and access patterns are often at the heart of performance issues in computing systems due to the use of certain hardware techniques to improve performance. Altering the access patterns to increase locality can dramatically increase performance of a given algorithm. Besides, repeated data access can be seen as redundancy in data movement. Similarly, there can also be redundancy in the repetition of calculations. This work also identifies some of the opportunities for avoiding these redundancies by directly reusing computation results. We document the possibilities of such reuse in some selected machine learning algorithms and give initial indicative results from our first experiments on data access improvement and algorithm redesign.

**KEYWORDS**

Increasing data locality, data redundancy and reuse, machine learning...

## 1. INTRODUCTION

Machine learning algorithms and their implementations typically are computationally intensive and also operate on large data sets. To learn, an ML algorithm needs to look at the training points (which together make up the training set). In supervised learning, the training set consists of data points for which the outcome to be learned is known: the points are labelled. Based on the training points, the algorithm builds a model by setting model parameters (also known as weights) so that the resulting model represents the underlying structure of the data. The model is evaluated on the test points to evaluate its accuracy. An algorithm and/or model may have extra parameters that are set by the user rather than learned from the data. These extra parameters are called hyperparameters. A particular algorithm being run on a particular data set to make a model is called a learner. To distinguish between particular hyperparameter settings for a learner we refer to a learner instance per hyperparameter tuple. The train and test steps are typically performed several times for different types of model and different model hyperparameters e.g. by dividing between training and test sets.

The reuse distance of a data location is the number of surrounding loop iterations that occur in between accesses to it. During the training phase, most machine learning algorithms have to read each data point in the training set at least once. Such a full traversal of the training set is called an epoch. Once a learner is trained, the resulting fixed model can be used in an operational testing phase to make predictions for previously unseen points. In the next sections we will investigate the potential for performance improvements. Unless otherwise stated, we are dealing with the training algorithms for a given learner rather than its use to classify unlabeled data in an operational environment. This is because the reuse available in the operational phase is often a) very little or zero and b) rather simple to exploit.

The remainder of this paper is as follows: in Section 2 an analysis of data and computational redundancy is given for the stochastic gradient descent along with experimental results. In Section 3, the instance based leaners class of models are presented together with empirical gains obtained when coupling these modes together. Section 4 deals with data reuse in logistic regression and support vector machines. We draw some conclusions is Section 5. The experiments presented in this paper have been conducted on a node from a 20 nodes cluster equipped with dual 6-core Intel(R) Westmere CPUs with 12 hardware threads each, a clock speed 2.80GHz and 96 GB of RAM. The implementations are all sequential (executed on one core) and C++ is used as a programming language. The results from the experiments are very much initial results given as an indication as to how some of the reuse and redundancy outlined in this document could be used.

## 2. RELATED WORK

To the best of our knowledge, no similar research work centered on analyzing data access patterns and identifying locality and redundancy for ML algorithms exist. Indeed, conventional contributions dealing with reducing data access overheads in ML suggest using distributed approaches [SHARED,CYCLADES]. In this case the focus is on single machine settings. We think however that such a review is a first step towards improving the performance and the efficiency in HPC settings.

## 3. STOCHASTIC GRADIENT DESCENT

The training or fitting of many ML models is based on optimisation techniques, with many approaches based on gradient based optimisation, using either first or second order gradients. Due to the expense of gradient calculations for complex models on large data sets, an approximation is often used, hence we consider these algorithms here. Stochastic Gradient Descent (SGD) is a quite popular algorithm. It is an optimization method that attempts to find the values of the model coefficients (the parameter or weight vector) that minimizes the loss function when they cannot be calculated analytically. SGD has proven to achieve state of-the-art performance on a variety of machine learning tasks [bottou_2012,sgdBPMF]. With its small memory footprint, robustness against noise and fast learning rates, SGD is indeed a good candidate for training data-intensive models. SGD is a variety of the gradient descent (GD) method [gradient_descent] with a major difference in the number of updates per visited data point. For SGD, only a random element from the training data is considered every iteration to perform the update of the weights vector. A template of both methods is described in *Algorithm 1* where the input parameter *n* defines the number of points to consider in the gradient computation and hence the algorithm. For SGD, *n* is equal to 1 while for GD the *n* is equal to the size of the complete training set.

```
Data: T, a training set
      M, an initial guess for the model parameters
      n, the batch size for calculating the gradient
for The required number of epochs do
    for Each batch B of size n in T do
        for Each training point t in B do
            Compute the gradient g of the loss function for M with respect to t
            Add g to the combined gradient G
        end
        Update model parameters M by a step in direction G
        Update algorithm parameters such as step size etc.
    end
end
```

*Algorithm 1: (Stochastic) Gradient descent algorithm template.*

The main data touched in both algorithms is the training set, the model and the calculated gradient. The reuse distance for any training point in both algorithms is $|T|$, the size of the training set. Similarly, any calculated gradient $g$ is used directly in the same iteration (reuse distance 0), and the model is reused every iteration (reuse distance 1). The learning rate parameter is updated as frequently as the weight vector for each algorithm.

Gradient descent-like optimization is often used with several learners. The data traversal and the number of data touches is largely determined by the optimization algorithm used *regardless of the model being trained*. Hence, one intuitive idea for a data reuse-aware coding is to fold different models together and train them simultaneously using the same optimization method, thus re-using the stream of data. A second approach we propose is a cache efficient SGD where the gradient is computed using new training points that have just been loaded from the large memory, along with others that are still in the cache memory.

## 2.1 CACHE-EFFICIENT SGD:

The basic idea of a cache efficient SGD is to also consider recently visited points in the computation of the gradient. The list of recently visited points is kept in a vector potentially saved in the cache memory. The technique relies on the fact that computing the differentiated loss function on larger sized batches that come from cache is almost a free operation compared to loading new training points from the main memory. Indeed, as an example of how effective caches can theoretically be, let us assume a simple computation where access to main memory takes 40 cycles and access to the cache memory take 4 cycles (such as on Intel(R) Westmere CPUs [Intel]). If the model uses 100 data elements 100 times each, the program spends 400,000 cycles on memory operations if there is no cache and only 40,000 cycles if all data can be cached. The principle of the sliding window can be applied to every level of the memory hierarchy though, in a nested fashion, thereby extending simple SW-SGD to cover more complex hierarchies. The application of SW-SGD could be complicated somewhat by the features of the cache that is being used to store the training points.

From machine learning and optimization prospective, using extra points in a gradient calculation should provide some extra smoothing similar to the extra points used in mini batch SGD (MB-GD). Ideally the smoothing would be as effective as that in MB-GD and achieve lower noise whilst having the same data touch efficiency as 1 point SGD. The cache efficient SGD does not introduce any parameters to the algorithm and it should be possible to apply the fundamental idea of this work to many SGD algorithmic variants.

The cache efficient SGD was experimented with a classification problem of the MNIST [MNIST] dataset containing 60,000 training and 10,000 testing images. It have also been tested on other gradient descent optimization algorithms [SGDvariant]such as Momentum, Adam, Adagrad, etc. The model to train is a neural network with 3 layers and 100 hidden units each. All the results are averaged from 5-fold cross-validation runs. A preliminary set of experiments was conducted in order to determine the best hyper-parameters (learning rate, batch size) of the algorithm. These best hyper-parameters have been used in a second experiments where the aim here to prove that using the data reuse optimization helps accelerating the convergence and that it is orthogonal to the other gradient algorithms. We experimented with three scenarios for every algorithm: (1) only a batch of $B$ new points ($B$ being the best batch size from the preliminary experiments), (2) $B$ new points + $B$ points from the previous iteration and (3) $B$ new points + $2 * B$ points from the previous iteration.

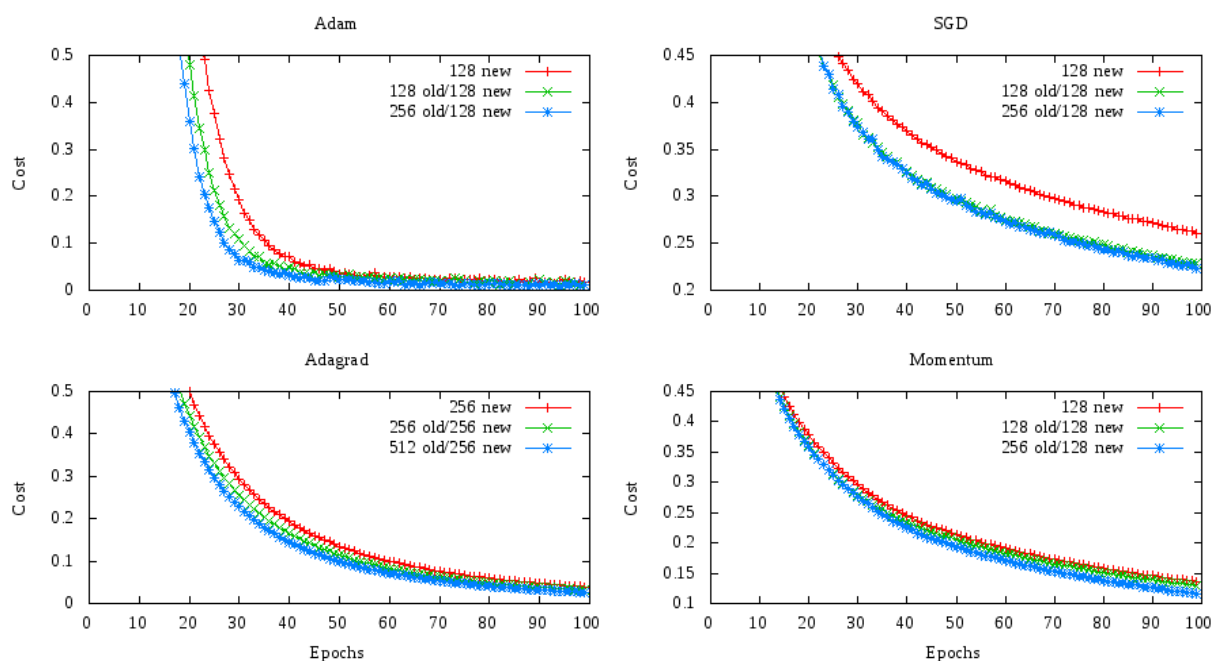

*Figure 2 Comparing different sizes of batches for different optimizers.*

In Figure 2, different sizes of batches are compared for different optimizers. For all algorithms, adding cached data points to the computation of the gradient improves the convergence rate. It is important here to highlight that, for the Adam algorithm for example, using a batch of size 256 and 512 new points is less efficient than using 128 new points as the first set of preliminary experiments showed that 128 is the best batch size. The added value here is therefore brought by the characteristic of considering old visited points in the computation and not because of a bigger batch size.

Further possible direction can be explored in this context such as loading different batches of points and to computing their gradients at the same time. Hence (1) future vectorisation opportunities are available, (2) the reuse distance is reduced for each point and as a consequence the temporal and spatial locality are increased.(3) matrix product best practices such as a cache blocking could be used where possible.

## 4. INSTANCE-BASED LEARNERS

In this section, we zoom in data and computation reuse possibilities in Instance-based class of learners (IBL), also known as memory-based learners. IBL is a class of ML algorithms that do not construct a model as in usual classification methods. They are called instance-based because they construct the hypotheses directly from the training points themselves, and the learner is ``memorising'' the training set. In several of these methods no training phase takes place (with the possible exception of a search for the best value of some hyperparameter) and the predictions are made by measuring similarities between points to predict and points from the training data set. The similarity measure and way of combining input from the training points defines the different types of instance-based learners.

## 4.1 K-NEAREST NEIGHBOURS:

The principle behind the $k$-nearest neighbour ($k$-NN) method is to find the $k$ training samples closest in distance to the new point, and predict the label from these. The distance can, in general, be any metric but the standard Euclidean distance is definitely the simplest and most used technique (if the attributes are all of the same scale). An unknown instance point is classified by a majority vote of its neighbours. For a classification problem, the output of *k*-NN is a class membership while for a regression problem, the output of *k*-NN is the average of the values of its *k* nearest neighbours. The algorithm template of the *k*-NN technique is given in Algorithm 2, but note that this is the classification phase, not the training phase as this instance based learner is not trained.

```
Data: RT, the set of remembered training points
      P, the set of points to predict
for all i points in P (1) do
    for all j remembered training points in RT (2) do
        d = compute_distance(i, j)
        if d ≤ the k closest distances then
            Add j to the list of k nearest neighbours of i
            (Remove further away members from the list if necessary)
        end
    end
    for all j in the list of k nearest neighbours of i do
        Add class of j to the running vote count for i
    end
    Return the class of i based on the majority vote
end
```

*Algorithm 2: Algorithm template of the k nearest neighbours method (classification).*

The data accessed here are the points in *RT* and the points in *P*. The point from *P* being classified is reused directly in each iteration of loop level 2, with a reuse distance of one. The reuse of training points from *RT* is carried by loop level 1, with reuse distance *RT*. Each execution of loop level 2 is an epoch. There is some reuse in the handling of the list of *k* nearest neighbours for any point, but this is likely to be a trivial amount of data and computation compared to handling *RT* and computing distances. The only simple way to improve reuse is to shorten the reuse distance for elements of *RT* by calculating distances to multiple prediction points simultaneously; an appropriate batch size can be calculated based on cache sizes available.

## 3.2 PARZEN-ROSENBLATT WINDOW

The second instance-based ML algorithm considered here is the Parzen-Rosenblatt window (PRW) density estimation method which has been introduced by Emanuel Parzen and Murray Rosenblatt respectively in [Parzen] and [Rosenblatt]. Its basic idea for predicting the probability density function P(x) is to place a window function f at x and determine what are the contribution of each training point $x_i$ to the window. The approximation to the probability density function value P(x) is computed using a kernel function that returns a weighted sum of the contributions from all the samples $x_i$ to this window. PRW has two important hyperparameters, the window bandwidth and the kernel function. A great number of bandwidth selection techniques exist such as [Jones1996, Sheather2004]. For the kernel function, different variants exist: Gaussian, Epanechnikov, Uniform, etc. The Gaussian probability density function is the most popular kernel for Parzen-window density estimation because it has no sharp limits as the kernels listed above, it considers all data-points and produces smooth results with smooth derivatives.

```
Data: RT, the set of remembered training points
P, the set of test points to predict
for all i test points in P (1) do
    for all j remembered training points in RT (2) do
        s = compute_similarity(i, j)
        Add s to the running total for the C, the class of j
    end
    Return the class of i based on the class with the highest total weight C
end
```

*Algorithm 3: Algorithm template of the Parzen-Rozenblatt Window method.*

From an implementation point of view, as sketched in the Algorithm template 3, computing PRW implies evaluating a large number of kernel-functions at a large number of data points. Given that the kernel function similarities are often based on Euclidean distance, the similarity with *k*-NN is obvious. The data reuse and distances involved are hence the same as for *k*-NN. The same optimizations are also applicable for the *k*-NN method as mentioned above.

### 3.4 COUPLING INSTANCE BASED LEARNERS:

Similar to other various machine learning algorithms, Parzen-Rozenblatt window and *k*-NN exhibit very similar data access pattern and share a large volume of the computations in different steps of the ML process: training, optimization, sampling, etc.

From a machine learning perspective, the Parzen window method can be regarded as a generalization of the *k*-nearest neighbour technique. Indeed, rather than choosing the *k* nearest neighbours of a test point and labelling the test point with the weighted majority of its neighbours' votes, one considers all points in the voting scheme and assigns their weight by means of the kernel similarity function. From a computation perspective, these algorithms similarly loop over all the points and sometimes calculate the same underlying distances (typically Euclidean). Therefore, the idea here is to run these two learners jointly on the same input data whilst producing different models. A proof of concept has been made on a subset of the Chembl public data set [CHEMBL] with 500K compounds and 2K targets. Our objective here was to give a first estimation of the amount of compute time that can be saved using the aforementioned optimization.

|  | Load time (s) | Test time (s) |
|---|---|---|
| PRW+*k*-NN separately | 7.545 | 2695.45 |
| PRW+*k*-NN jointly | 3.726 | 1601.035 |

*Table 1: Comparing the elapsed time when running PRW and k-NN separately and jointly.*

In Table 1, the results are represented for two scenarios: both learners are run separately and the total elapsed time is accumulated and both learners run jointly. The elapsed times are in seconds. Two steps are considered: the time for loading the training and testing sets and the time for the testing step (recall here that in instance-based learners no actual training phase occurs).

The preliminary experimental results clearly show the added value of increasing data reuse when computing PRW and k-NN on the same pass over the data. The computing time is indeed almost divided by two.

### 5. LOGISTIC REGRESSION AND SUPPORT VECTOR MACHINE

Logistic Regression (LR) and Support Vector Machine (SVM) are two closely related approaches for fitting linear models for binary classification. The optimum placement of the hyperplane dividing the two classes is determined by the loss function, which determines the per training point contribution to the total cost of any given model parameters, or the gradient of the cost which can then be used for optimisation. Optimisation is made easier by the fact that the loss function is convex.

In terms of the access to data, the two approaches are essentially the same. For a given batch of training points, the distance of the point to the model hyperplane is calculated with an inner-product per training point to produce a scalar, which is then fed into the differentiated loss function (to calculate the gradient). For the purpose of discussion we assume an Euclidean distance and the associated inner-product. Algorithm 4 describes the procedure (the bias term is omitted for clarity).

```
Data: A batch of training samples B
An initial guess for the hyperplane model M
for all training points t in B (1a) do
    for all vector indices i in t and M (2) do
        Multiply scalars t_i and M_i
        Accumulate in p (the inner product)
    end
    Accumulate the gradient f'(p) in g (the batch gradient)
end
for all weights in model M (1b) do
    update the weight based on weight decay, training schedule step size and the entry in g
end
Return M
```

*Algorthim 4: Pseudo-code for updating a linear model with a mini-batch gradient descent.*

The reuse in a single batch update for these algorithms is relatively limited. Each training point $t$ in the batch is accessed only once. The majority of accesses to the model $M$ is carried by loop (1a); there is one inner-product per training point. This leads to a reuse distance of $|M|$. These loops can be easily rearranged to carry out the inner-products simultaneously so that there is one traversal over the model for those operations. The update of the weights in loop (1b) is, in the basic algorithm, dependent on the gradient calculated in loop (1a), so these two calculations are effectively sequentialised.

If these two algorithms are to be run on the same training set note that they can be quite tightly coupled. The training points can be presented to both models in the same order, meaning that the common visit to a training point can be combined so that the data only needs to be touched once. Indeed, the inner-product of the training point with the different hyperplane models can be done at the same time so that there is direct reuse in a feature-by-feature way of the training point.

## 4.1 JOINTLY TRAINING LOGISTIC REGRESSION AND SVM:

In this section, an experimental investigation is conducted to explore the impact of exploiting the data reuse opportunities related to logistic regression and the support vector machine collectively trained by SGD. Note that the algorithms are not being changed in any way, this is pure leveraging of available reuse. Recall that LR and SVM are very closely related approaches in terms of data accesses especially when they are trained by optimization algorithms based on gradient calculation. In this case, only the computation of the cost functions are different. However, these two functions are applied in the exact same order and at the same time during the model training. By training two learners we are adding an outer loop over them to Algorithm 4. We then perform interchange of the mini-batch loop and the learner loop to be able to exploit reuse.

We have empirically compared the impact of training LR and SVM using the SGD optimization algorithm on two different sized benchmarks also used in state of the art related work [bottou2010]. The benchmark dataset is extracted from the Pascal challenge on large scale hierarchical classification [pascal_dataset] which is a text classification challenge. The ALPHA data set is composed of 250,000 training point and 250,000 test points.

The DNA data set is composed of 4,038,390 train points and 2,500,000 test points. In Table 2, we report the elapsed time in seconds for running SVM and LR separately and jointly on the aforementioned datasets.

|  | Alpha dataset | | DNA dataset | |
| --- | --- | --- | --- | --- |
|  | Train time (s) | Test time (s) | Train time (s) | Test time (s) |
| SVM + LR separately | 49.9833 | 24.6442 | 338.559 | 155.2869 |
| SVM + LR jointly | 45.6416 | 22.0608 | 279.708 | 115.644 |

*Table 2: Comparing the elapsed time when running LR and SVM separately and jointly.*

The obtained results show that reducing the data traversal decreases the total execution time and that the gain increases with the size of the data used. Indeed, with the DNA dataset the total training and testing time is reduced by around 20% compared to the time measured when the two models are run separately more than the time gained with the alpha dataset. It is important here to highlight that the output predictions are exactly the same with the two scenario because the essence of the models are not modified. It should also be noted that these are very early results, and that the benefit of combining the models is not yet as large as initially expected.

# 6. CONCLUSION:

In this research paper we have highlighted various access patterns with reuse and computations containing redundancy for a set of ML algorithms that we think are most used and useful. For each of these we investigated the potential for performance improvements by identifying the reuse distances for the data used and the computations that could be grouped mainly between different learners. This work is intended as a starting point for further work on improving the performance of learners as implemented on supercomputing clusters and applied to big data volumes.

# ACKNOWLEDGMENTS

The research leading to these results has received funding from the European Union's Horizon2020 research and innovation programme under the EPEEC project, grant agreement No 801051.

The final publication is available in the proceedings of the "IADIS International Conference Big Data Analytics, Data Mining and Computational Intelligence 2019 (part of MCCSIS 2019), at http://www.iadisportal.org/digital-library/reviewing-data-access-patterns-and-computational-redundancy-for-machine-learning-algorithms

# 7. REFERENCES

[Parzen] Parzen, Emanuel. On Estimation of a Probability Density Function and Mode. The Annals of Mathematical Statistics 33 (1962), no. 3, 1065–1076.

[Rosenblatt] Rosenblatt, Murray. Remarks on Some Nonparametric Estimates of a Density Function. The Annals of Mathematical Statistics 27 (1956), no. 3, 832–837.

[bottou_2012]. Bottou, Léon. Stochastic Gradient Tricks. Neural Networks, Tricks of the Trade, Reloaded , pages 430-445, 2012.

[bottou_2010] Bottou, Léon. Large-Scale Machine Learning with Stochastic Gradient Descent. Proceedings of the 19th International Conference on Computational Statistics (COMPSTAT'2010). Paris, France, 2010, pages 177-187.

[sgdBPMF] S. Ahn et al. Large-scale distributed bayesian matrix factorization using stochastic gradient MCMC. Proceedings of the 21th ACM SIGKDD International Conference on Knowledge Discovery and Data Mining, 2015.

[CHEMBL] A. Bento et al. The ChEMBL bioactivity database: an update. Nucleic Acids Res., vol. 42, pp. 1083–1090, 2014.

[SHARED] S. Sallinen et al. "High Performance Parallel Stochastic Gradient Descent in Shared Memory," 2016 IEEE International Parallel and Distributed Processing Symposium (IPDPS), Chicago, IL, 2016, pp. 873-882.

[CYCLADES] P. Xinghao et al. "CYCLADES: Conflict-free Asynchronous Machine Learning" . eprint arXiv:1605.09721 05/2016.

[gradient_descent] https://en.wikipedia.org/wiki/Gradient\_descent.

[Intel] http://www.7-cpu.com/cpu/Westmere.html

[Jones1996] Jones, M et al. A brief survey of bandwidth selection for density estimation, Journal of American Statistical Association, vol 91, 401-407, 1996.

[Sheather2004] Sheather, S. J. Density estimation. Statistical Science, 19, 588–597, 2004.

[pascal_dataset] http://lshtc.iit.demokritos.gr/node/1.

[MNIST] LeCun, Y et al. (1998). Gradient-based learning applied to document recognition. Proceedings of the IEEE, 86, 2278-2324.

[SGDvariant]http://sebastianruder.com/optimizing-gradient-descent/index.html#gradientdescentoptimizationalgorithms

[IDA] Chakroun, Imen & Vander Aa, Tom & Ashby, Tomas. (2019). Guidelines for enhancing data locality in selected machine learning algorithms. Intelligent Data Analysis. 23. 1003-1020. 10.3233/IDA-184287.